\documentclass[runningheads]{llncs}
\usepackage{graphicx}
\usepackage{tikz}
\usepackage{comment}
\usepackage{amsmath,amssymb} % define this before the line numbering.
\usepackage{color}
\usepackage{multirow}
\usepackage{booktabs}
\usepackage{algorithm}
\usepackage{algpseudocode}
\usepackage{utfsym}
\def\dui{\textcolor[rgb]{0,1,0}{\usym{2714}}}
\def\cuo{\textcolor[rgb]{1,0,0}{\usym{2718}}}
\usepackage[width=122mm,left=12mm,paperwidth=146mm,height=193mm,top=12mm,paperheight=217mm]{geometry}

\usepackage[pagebackref=true,breaklinks=true,colorlinks,bookmarks=false]{hyperref}
\usepackage{xcolor}
\usepackage{colortbl}

\makeatletter
\renewcommand*{\@fnsymbol}[1]{\ifcase#1\or*\else$\dagger$\fi}
\makeatother

\makeatletter
\newcommand{\printfnsymbol}[1]{%
  \textsuperscript{\@fnsymbol{#1}}%
}
\makeatother

\begin{document}
\pagestyle{headings}
\mainmatter
\def\ECCVSubNumber{4338}  % Insert your submission number here

\title{TransGrasp: Grasp Pose Estimation of a\\Category of Objects by Transferring Grasps\\from Only One Labeled Instance} % 
\titlerunning{TransGrasp}
\author{Hongtao Wen\thanks{H. Wen and J. Yan—Equal contributions.}\and
Jianhang Yan\printfnsymbol{1} \and
Wanli Peng\thanks{Corresponding author.} \and
Yi Sun
}
\authorrunning{H. Wen et al.}
\institute{Dalian University of Technology, China\\
\email{\{wenht,yjh97,1136558142\}@mail.dlut.edu.cn}, \email{lslwf@dlut.edu.cn}
}
\maketitle
\begin{abstract}
Grasp pose estimation is an important issue for robots to interact with the real world. However, most of existing methods require exact 3D object models available beforehand or a large amount of grasp annotations for training. To avoid these problems, we propose TransGrasp, a category-level grasp pose estimation method that predicts grasp poses of a category of objects by labeling only one object instance. Specifically, we perform grasp pose transfer across a category of objects based on their shape correspondences and propose a grasp pose refinement module to further fine-tune grasp pose of grippers so as to ensure successful grasps. Experiments demonstrate the effectiveness of our method on achieving high-quality grasps with the transferred grasp poses.
Our code is available at
\href{https://github.com/yanjh97/TransGrasp}{https://github.com/yanjh97/TransGrasp}.
\end{abstract}

\section{Introduction}
\label{sec:intro}

\begin{figure}[t]
    \centering
		\includegraphics[width=0.9\linewidth]{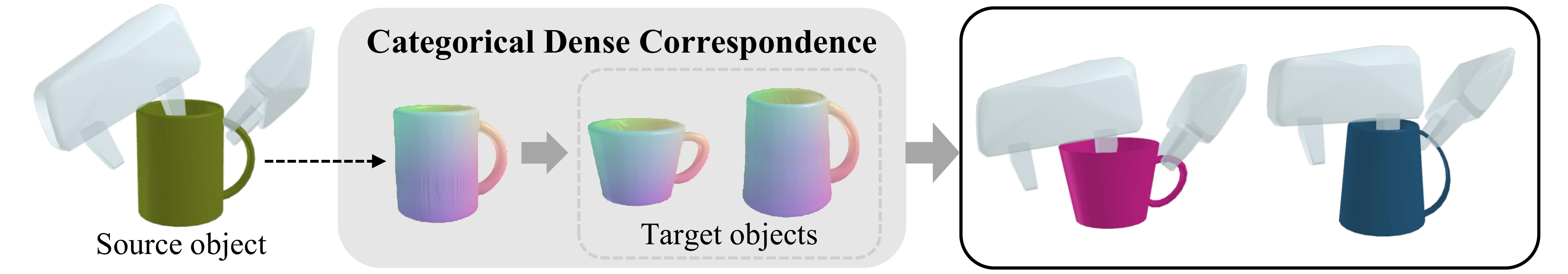}
	\setlength{\abovecaptionskip}{-0.1em}
	\caption{
		Our method for Grasp Pose Estimation with \textbf{only one labeled instance}.
		We transfer grasps from one labeled source object to target objects, using the learned categorical dense correspondence among a category of objects
	}
	\label{fig:teaser}
\end{figure}

Grasp pose estimation refers to the problem of finding a grasp configuration for the grasping task given an object. It's one of the most important issues for robots to interact with the real world and has great potential to replace human hands to execute various tasks, such as helping elderly or disabled people carry out everyday tasks. However, predicting a stable grasp of an object is a challenging problem as it requires modeling physical contact between a robot hand and various objects in real scenarios.
Currently, most existing works rely on grasp pose annotations on each object or exact 3D object models available beforehand.
However, annotating 6D grasp pose is not easy as the rotation, translation and width of the gripper have to be adjusted to ensure a stable grasp.
In this paper, we deal with the problems from the perspective of computer vision to develop a category-level grasp pose prediction model. Our approach aims to estimate the grasp poses of all instances by labeling the grasp poses on only one instance of the category and transferring the grasp to other instances, as shown in Fig.~\ref{fig:teaser}. As we have known, the study on category-level grasp pose estimation has not been fully explored so far.

Generally, grasp pose estimation can be divided into model-based methods \cite{miller2004graspit,leon2010opengrasp,zeng2017multi,xiang2017posecnn,wang2019densefusion,he2020pvn3d} and data-driven (model-free) methods \cite{ten2017grasp,liang2019pointnetgpd,mousavian20196,murali20206,qin2020s4g,fang2020graspnet,breyer2021volumetric,wu2020grasp}. Both methods have their own limitations. Model-based methods generate grasp poses of an object regarding the criteria such as force closure. It requires precise geometric and complete 3D model of an object which is not always available in real scenarios, while data-driven methods estimate grasp poses by experience that is gathered during grasp execution. It requires dense grasp pose annotations which are extremely time-consuming and is also uncertain whether the experience can be generalized to novel objects. In this paper, we propose a novel grasp pose estimation approach, named TransGrasp, to solve the above problems. 

Our TransGrasp is based on the fact that a category of object instances have consistent topologies and grasping affordances, such as mugs with grasping handles for drinking. This motivates us to align grasp contact points within a category of objects by first building their shape correspondences and then transfer labeled grasps on one object to the others, finally these initial grasps on all objects are fine-tuned continuously. In this work, we first learn the categorical shape latent space and a template that captures common structure of the category from the given object instances. The variance of each object instance is represented by the difference of its shape relative to the template, and dense correspondence across instance shapes can be established by deforming their surfaces to the template. Then we design a novel model to transfer grasp points on one object instance across a category of objects based on their shape deformation correspondence. To avoid extreme distortions of grasp points during correspondence deformation, the grasp pose refinement module is introduced to further alter the gripper, allowing the grasp points on the surface to form a stable grasp. Note that the correction on gripper is crucial to guarantee a successful grasp. Finally, grasp points of gripper on any object surface can be aligned across an entire category of objects. 

Compared with previous works, the proposed model predicts grasp poses of an entire category of objects with grasp pose annotations of only one object instance, which greatly reduces the amount of 6DoF grasp pose annotations of objects in data-driven methods. Furthermore, category-level grasp pose estimation model learns a categorical shape latent space shared by all instances, which enables shape reconstruction and grasp pose estimation within a category. We additionally estimate rotation $R$, translation $t$, and scale $s$ given object observable point cloud as input to transform grasp poses from Object Coordinates to Camera Coordinates and subsequently conduct simulation and real robot experiments. The experiment results verify that our estimated grasp poses can align well across the instances of a category.

The contributions of this work are as follows:

$\bullet$ We propose a category-level grasp pose estimation model to predict grasp poses of a category of objects by only labeling one object instance, which does not heavily relying on a large amount of grasp pose annotations and 3D model on every object.

$\bullet$ We exploit consistent structure within a category of objects to perform grasp pose transfer across a category of objects. In addition, the grasp pose refinement module further corrects grasp poses to ensure successful grasps.

$\bullet$ We set up simulated and real robot system and conduct sufficient experiments that verify the effectiveness of the proposed method on achieving high-quality grasps with the transferred grasp poses.

\section{Related Work}
\label{sec:related}
In this section, we briefly review the literatures on 6DoF grasp pose estimation, which can be approximately divided into model-based methods and data-driven (model-free) methods. 

\noindent \textbf{Model-based methods:}
A common pipeline for model-based methods is to generate the grasp pose on the known object models using the criteria such as force closure \cite{miller2004graspit,leon2010opengrasp}, and then transform annotated grasps from Object Coordinates to Camera Coordinates with the estimated object 6D pose. Zeng \emph{et al.} \cite{zeng2017multi} segment objects with fully convolutional network from multiple views and fit pre-scanned 3D models with ICP \cite{besl1992method} to obtain object pose. Others apply learning-based and instance-level object pose estimation methods such as \cite{he2020pvn3d,wang2019densefusion,xiang2017posecnn} for grasp transformation. However, these model-based methods require exact 3D models beforehand, which cannot be used in general settings because the target objects may be different from the existing 3D models. To estimate the 6D pose of unseen objects, researches for category-level object pose estimation \cite{Wang_2019_CVPR,chen2020learning,tian2020shape,chen2021fs} have drawn more and more attention, but they have hardly been applied to the grasp estimation. In this work, we explore the challenging problem for category-level grasp pose estimation by transferring grasps from one labeled instance to the others.

\noindent \textbf{Model-free methods:}
To estimate the grasp pose on unseen objects, some model-free methods \cite{ten2017grasp,liang2019pointnetgpd,mousavian20196,murali20206,qin2020s4g,fang2020graspnet,breyer2021volumetric,wu2020grasp} that analyze the 3D geometry of the object surface directly using deep neural network are proposed. GPD \cite{ten2017grasp} and PointNetGPD \cite{liang2019pointnetgpd} generate a large set of grasp candidates firstly and then build a CNN or PointNet \cite{qi2017pointnet} to identify whether each candidate is a grasp or not. Following this pipeline, 6-DOF GraspNet \cite{mousavian20196} builds a variational auto-encoder (VAE) for efficiently candidates sampling and introduces a grasp evaluator network for grasp refinement. CollisionNet \cite{murali20206} builds a collision evaluation extension based on \cite{mousavian20196} to check the cluttered environment for collisions. Since analyzing every candidate costs significantly, some end-to-end methods \cite{qin2020s4g,fang2020graspnet,breyer2021volumetric,wu2020grasp} are proposed. Among them, S$^4$G \cite{qin2020s4g} builds a single-shot grasp proposal network based on PointNet++ \cite{qi2017pointnet++} to predict 6DoF grasp pose and quality score of each point directly. Fang \emph{et al.} \cite{fang2020graspnet} build a large scale grasp pose dataset called GraspNet-1Billion and propose an 3D grasp detection network that learns approaching direction, operation parameters and grasp robustness end-to-end. However, all these methods need a sufficient amount of dense grasp pose annotations for all the objects in the scene for training, which is time-consuming and laborious. To avoid this problem, our method estimates grasp poses of a category of objects by labeling only one object instance, which greatly reduces the cost of grasp pose annotations.

Closely related to our work, some methods transfer grasps to other objects by building dense correspondence. DGCM-Net \cite{patten2020dgcm} transfers grasps by predicting view-dependent normalized object coordinate (VD-NOC) values between pairs of depth images. CaTGrasp \cite{wen2021catgrasp} maps the input point cloud to a Non-Uniform Normalized Object Coordinate Space (NUNOCS) where the spatial correspondence is built. DON \cite{florence2018dense} learns pixel-wise descriptors in a self-supervised manner to transfer grasp points between 2D images. NDF \cite{simeonov2021neural} builds 3D dense correspondence using Neural Descriptor Field, through which the grasp poses are transferred directly. Different from these methods, we construct dense surface correspondence between the complete shape of objects to transfer the grasp points on the surface to obtain coarse grasps, and further refine these grasps using a refinement module to ensure reliable and successful grasps for the gripper.

\section{Method}
Our ultimate goal is to estimate grasp poses of a category of objects by labeling only one object instance, rather than heavily relying on a large amount of grasp pose annotations on every object. Our TransGrasp establishes point-to-point dense correspondence among intra-class shapes (Sec.~\ref{sec:learnlatent}) to transfer grasp points on one source object instance to other target object instances by aligning grasp poses on the object surface (Sec.~\ref{sec:posegen}). Furthermore, to avoid extreme distortions during grasp transfer, we introduce a grasp pose refinement module to further adjust the pose of gripper, encouraging the grasps to satisfy antipodal principle (Sec.~\ref{sec:refinement}). Our overall architecture is illustrated in Fig.~\ref{fig:overview}. In the remainder of this section, we will discuss each component in detail.

\subsection{Learning Categorical Dense Correspondence}
\label{sec:learnlatent}
\begin{figure}[t]
    \centering
		\includegraphics[width=0.97\linewidth]{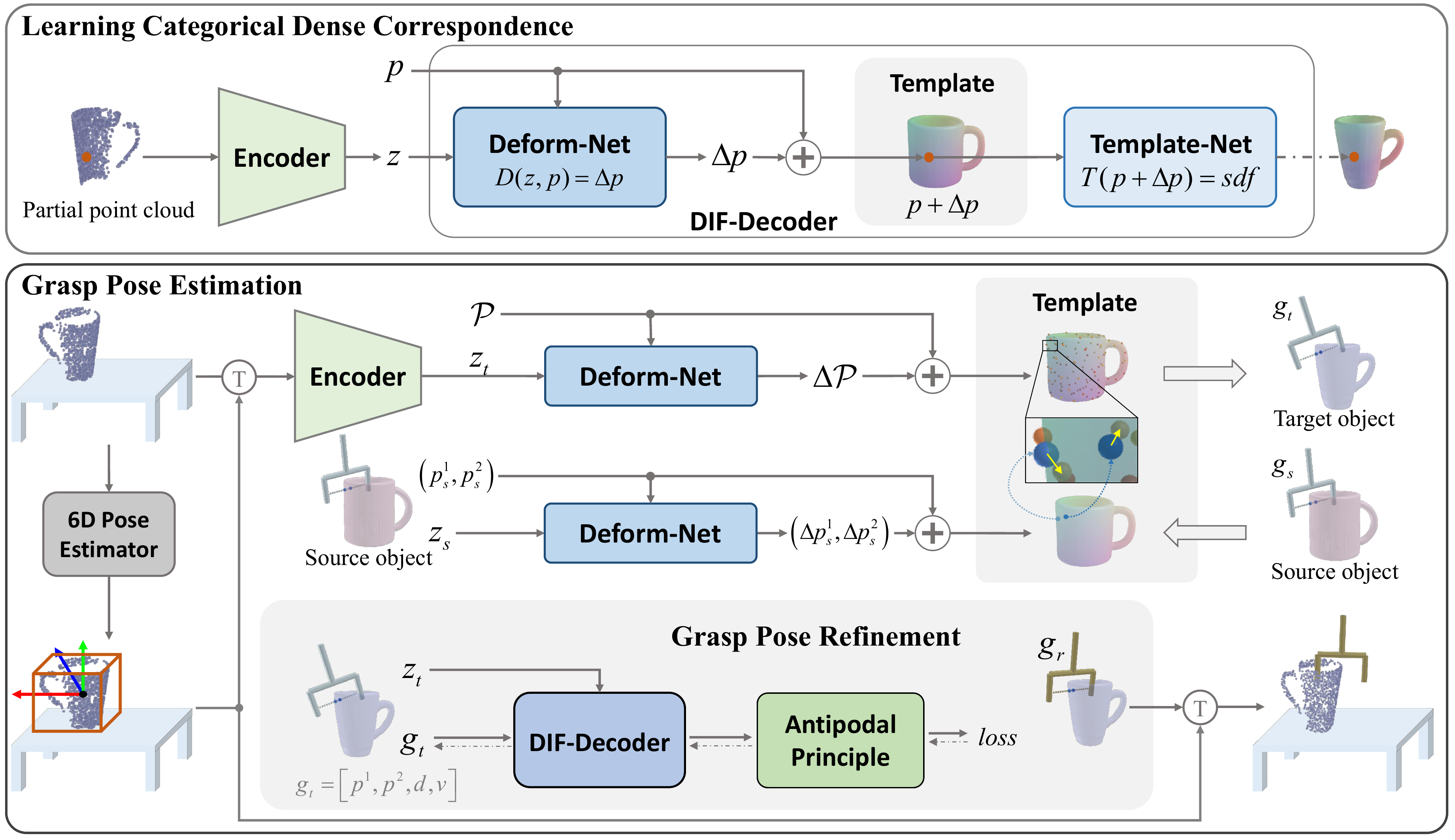}
	\setlength{\abovecaptionskip}{-0.5em}
	\caption{
		\textbf{Overview of the proposed TransGrasp.}
		Top: The network composed of Shape Encoder and DIF-Decoder is trained to learn categorical dense correspondence by deforming object points to the template.
		Bottom: During inference process, point clouds are first transformed from Camera Coordinates to Object Coordinates and then serve as the input of our trained network, generating the object instance's deformation to template. Meanwhile, the source model with grasp annotation $g_{s}$ is fed into Deform-Net to obtain its deformation. The dense correspondence established by their common template subsequently guides the transfer from $g_{s}$ to $g_t$ by aligning grasp points on object surface. The transferred grasp pose $g_t$ is then refined to $g_{r}$ by the grasp pose refinement module. Finally, $g_{r}$ is transformed from Object Coordinates to Camera Coordinates so as to conduct a grasp
	}
	\label{fig:overview}
\end{figure}

Objects can be classified according to shape, function, \emph{etc}. Objects classified by shape usually have consistent topologies and grasping affordances and they are likely to be successfully grasped in a similar way. Based on this prior, we can build the categorical dense correspondence across object instances so that grasp poses could be transferred from one object to others. The categorical correspondence across object instances needs to handle the intra-class shape variation and most recent works model the variation explicitly from a pre-learned template \cite{zheng2021deep,tian2020shape,deng2021deformed}. The variance of each object instance is represented by the difference of its shape relative to the template, and dense correspondence among instance shapes can be established through the template. In this paper, we build an encoder-decoder framework to reconstruct object model given partial point cloud and provide dense shape correspondence between the reconstructed model and the template.
The encoder-decoder framework is illustrated at the top of Fig.~\ref{fig:overview}.

Firstly, a latent shape space representing various 3D shapes and a template of a class for shape correspondence are learned by a decoder-only network.
Here we adopt Deformed Implicit Field \cite{deng2021deformed} decoder (DIF-Decoder), with a continuous random variable $z$ as input to adapt to the intra-class shape variation. The DIF-Decoder can be divided into two parts, a Template-Net ($T$) and a Deform-Net ($D$). The Template-Net can map a 3D point $p$ to a scalar SDF value $sdf$:
\begin{equation}
T (p) = sdf, \; p \in \mathbb{R}^{3}, \; sdf \in \mathbb{R}.
% T: p_T\in \mathbb{R}^{3} \to sdf\in \mathbb{R}.
\label{eq:dif_o_6}
\end{equation}
The network weights of $T$ are shared across the whole class, therefore it is enforced to learn common patterns within the class.
For each object, the Deform-Net learns a deformation offset to the template $D (z,p) = \Delta p$ for each query point $p$ with the latent code $z$. And then, the SDF value of point $p$ can be obtained via Template-Net $T (p + \Delta p) = sdf$.
This process establishes dense correspondence between an object instance and the template. In other words, the implicit template is ``warped'' according to the latent codes to model different SDFs. Therefore, dense correspondence among instance shapes can be established through the template. This provides a good reference for grasp transfer across object instances. The DIF-Decoder is trained separately for each category using the CAD models selected from ShapeNetCore \cite{shapenet2015}. After training, the Template-Net captures the common properties of one certain category and Deform-Net establishes the dense correspondence between instances and the template. Meanwhile, each object instance is mapped to a low-dimensional shape vector $z$. 

However, input data are usually partial point clouds in practice. To complete the shape information, a Shape Encoder learns to map each input to a latent shape code $z$ so that given the shape code a complete shape can be reconstructed using the priors learned by the DIF-Decoder. The Shape Encoder is trained by taking a partial point cloud rendered from its CAD model as input and shape vector $z$ of each object instance as output.

\subsection{Grasp Pose Estimation}
\label{sec:posegen}
After learning the categorical dense correspondence among different objects, we then focus on estimating grasp poses for every object instance by grasp pose transfer from only one labeled object. This transfer process is illustrated in the middle of Fig.~\ref{fig:overview}. The source object is the only one object instance with grasp pose annotations, and its grasp poses on the object surface will be transferred to other target object instances by its established correspondence to other ones. We first introduce a novel grasp pose representation that establishes the connection between grasp pose of the gripper and object surface points, and then implement the grasp pose transfer from the source object surface to the target object surface. 

\begin{figure}[t]
    \centering
		\includegraphics[width=0.86\linewidth]{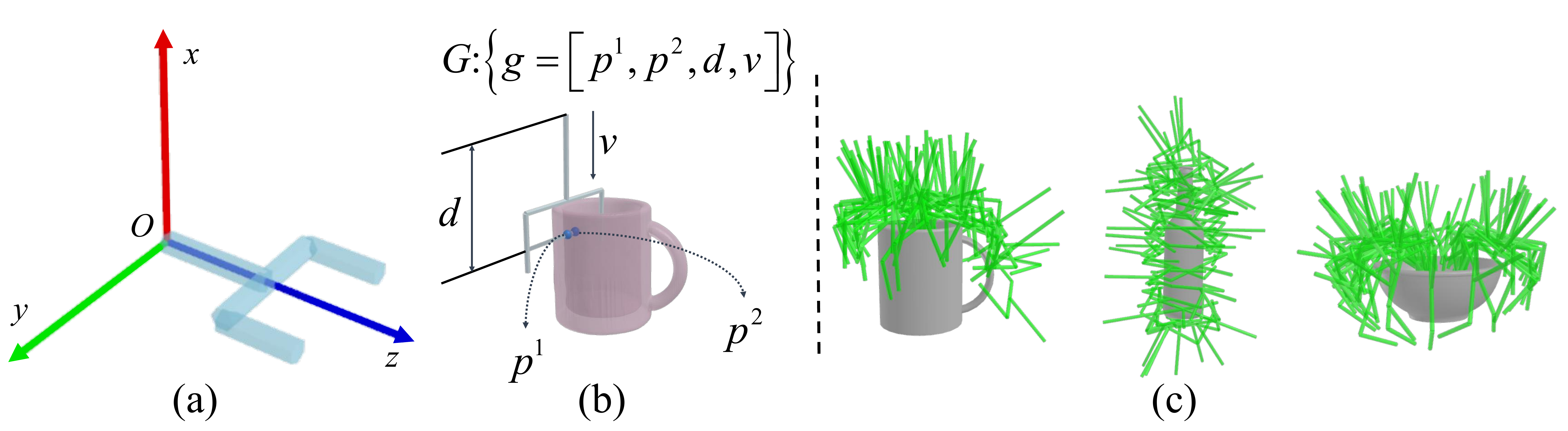}
	\setlength{\abovecaptionskip}{-0.6em}
	\caption{
	Grasp pose representation and grasp annotations on source object.
	(a) two-finger gripper's mesh model; (b) grasp representation where ($p^1$, $p^2$) are the grasp points on the object surface projected from gripper, with $d$ and $v$ denoting approaching depth and approaching vector, respectively; (c) The visualization of 50 grasp annotations on the source object for each category
	}
	\label{fig:method_grasppose}
\end{figure}

\noindent
\textbf{Grasp Pose Representation:}
To easily implement grasp transfer using dense correspondence between objects, we seek a novel suitable grasp representation for the gripper which follows the two-finger gripper's mesh model shown in Fig.~\ref{fig:method_grasppose}(a). Unlike previous works \cite{mousavian20196,murali20206,yang2021robotic} that define the grasp pose as ($R$, $t$, $width$) of the gripper, we introduce a novel grasp representation which connects grasp pose with the object surface points as illustrated in Fig.~\ref{fig:method_grasppose}(b). Formally, we denote the two grasp points on the object surface projected from gripper as $p^1$, $p^2$, the vertical distance from gripper's origin $O$ to the line linking  $p^1$ and $p^2$ as approaching depth $d$, and the approaching vector as $v$. These four parameters form the grasp pose of the gripper where $v$ determines the approaching direction and $d$ the approaching depth along this direction. The width of gripper is determined by the distance between the two grasp points. In this section of Grasp Pose Estimation, only grasp points ($p^1$, $p^2$) are transferred from the labeled source object to other target objects keeping the $d$ and $v$ unchanged, while in the next section of Grasp Pose Refinement (Sec.~\ref{sec:refinement}) all pose parameters ($p^1$, $p^2$, $d$, $v$) are simultaneously refined to obtain optimal stable grasp poses.

\noindent
\textbf{Labeling One Instance:}
To realize grasp transfer among the instances of a class, it is essential to provide a known instance with grasp pose annotations as the source object. Therefore, for each category, one of the instances needs to be labeled in advance. For convenience, we randomly select one instance for each category from the existing large-scale grasp dataset ACRONYM \cite{acronym2020} with pose annotations. Meanwhile, to ensure their robustness and avoid inaccuracy, all the chosen grasps are tested in the simulation environment IsaacGym \cite{makoviychuk2021isaac}, and only the grasps that can successfully grasp the corresponding source objects in the simulation platform are regarded as the source for grasp transfer. We have about 1k grasp poses for each categorical source object. Figure~\ref{fig:method_grasppose}(c) shows 50 grasp annotations on the source object for each category.

\noindent
\textbf{Grasp Transfer:}
With our grasp pose representation and one labeled source object, the grasp poses could then be transferred to other object instances of the same category during the inference procedure of our approach shown in the center of Fig.~\ref{fig:overview}. As the encoder and decoder have been learned previously in Sec.~\ref{sec:learnlatent}, the shape dense correspondence between the source and target object can be established through the template immediately. Given grasp points $p^1_s$, $p^2_s$ on the source object, their deformation offsets $\Delta p^1_s$,  $\Delta p^2_s$ to the template are predicted by Deform-Net, then its grasp poses on the template will be obtained by this deformation accordingly. 
To further transfer the grasp points on the template to any target object, the input partial point cloud of the target object is first transformed from Camera Coordinates to Object Coordinates using off-the-shelf 6D object pose estimator such as \cite{tian2020shape}, then encoded to the latent code $z_t$ from which its complete shape is reconstructed using DIF-Decoder. And then query point set $\mathcal{P}$ on the surface of reconstructed model is deformed to the template by Deform-Net. Finally, the grasp poses on the template transferred from source object are naturally aligned to the target object through their shared template. It is noted that although the deformed source and target object shape have similar shapes on the template, the grasp points on the source object cannot be accurately matched to surface points on the target.
Here we apply the nearest point matching to transfer the grasp points on the source (\textcolor[RGB]{59,99,158}{blue points}) to the target (\textcolor[RGB]{190,143,0}{gold points}), as shown in the enlarged detail in Fig.~\ref{fig:overview}.

\subsection{Grasp Pose Refinement}
\label{sec:refinement}
Although point-to-point dense correspondence can realize the grasp transfer from a source object with grasp pose annotations to any target object, when handling some outlier objects with strange shapes, poor deformation could be estimated by Deform-Net, which in turn leads to the failure of grasp transfer. Therefore, we propose a grasp pose refinement module to improve performance after grasp transfer, as shown at the bottom of Fig.~\ref{fig:overview}.
To refine grasp pose, we utilize spatial relationship between gripper and object by calculating the distances from the points on gripper to the object surface. Specifically, the transferred grasp points $(p^{1}, p^{2})$ and the sampled points $p^{g}$ from the gripper, along with the latent code of target object $z_t$, are input to the DIF-Decoder to get the SDF values of these points.
The grasp pose is iteratively refined following the refinement principles for parallel-jaw gripper, until a stable grasp pose $g_r$ is achieved. The gripper poses before and after adjustment are shown as the silver and gold grippers in Fig.~\ref{fig:overview} respectively.

\begin{figure}[t]
    \centering
	\includegraphics[width=0.75\linewidth]{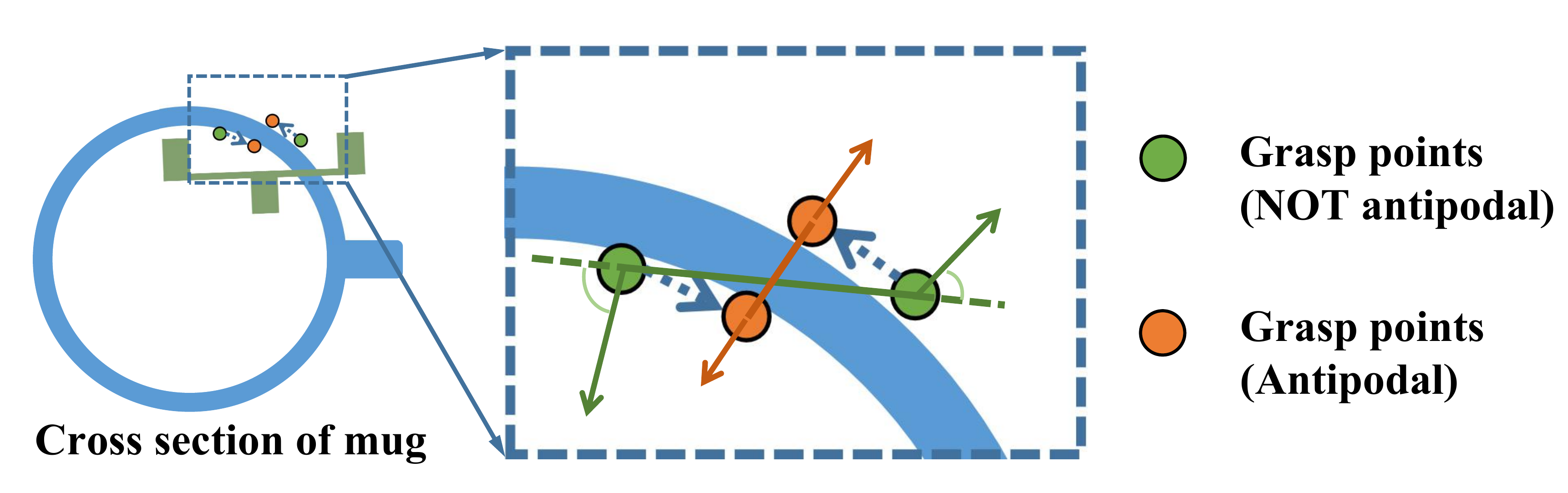}
	\setlength{\abovecaptionskip}{-0.8em}
	\caption{
		The refinement process of grasp points based on antipodal principle
	}
	\label{fig:method_refine}
\end{figure}

We first encourage the two grasp points to satisfy the antipodal principle (a simplified version of force closure for parallel-jaw gripper) on the object surface, as shown in Fig.~\ref{fig:method_refine}. In mathematics, antipodal points of a sphere are those diametrically opposite to each other. For the two-finger grasping problem, the grasp points are antipodal when their normals are in opposite directions and stay in one line with the grasp points. Let $n^1$ and $n^2$ be the normal vectors at $p^1$ and $p^2$ respectively, then they need to satisfy:
\begin{equation}
\cos \left \langle p^2 - p^1, n^1 \right \rangle = 1,
\cos \left \langle p^1 - p^2, n^2 \right \rangle = 1,
\label{eq:antipodal1}
\end{equation}
where $n^1$ and $n^2$ can be easily computed by taking the derivative of $sdf$ with respect to the grasp point: $\partial sdf/\partial p$, using the back-propagation of DIF-Decoder.

Therefore, we define the antipodal loss that encourages the grasp points to satisfy the antipodal principle:
\begin{equation}
\begin{aligned}
L_{anti} = -\cos \left \langle p^{2} - p^{1}, n^{1} \right \rangle-\cos \left \langle p^{1} - p^{2}, n^{2} \right \rangle.
\label{eq:loss_anti}
\end{aligned}
\end{equation}

In addition, we constrain the SDF value of the grasp points $(p^1$, $p^2)$ to zero, making sure that the grasp points are on the object surface using the touch loss:
\begin{equation}
L_{touch} = \Vert sdf^{1} \Vert_2 + \Vert sdf^{2} \Vert_2,
\label{eq:loss_touch}
\end{equation}
where $sdf^{1}$ and $sdf^{2}$ are SDF values of the transferred two grasp points $(p^1$, $p^2)$.

Moreover, to penalize the interpenetration between the gripper and object, we build a collision avoidance loss to constrain the gripper outside the object, that is, the SDF values of the gripper points are greater than 0 via:
\begin{equation}
L_{collision} = -\frac{1}{N} \sum_{i}^{N} \min (0,sdf^{g}_{i}),
\label{eq:loss_collision}
\end{equation}
where $N$ is the number of the sampled points $p^{g}$ of the gripper at its grasp pose, and $sdf^{g}$ are the SDF values of the these sampled points. Based on such a loss function, the sampled points of the gripper inside the object, of which the SDF values are less than 0, are pushed out of the object surface, so as to avoid the collision between the gripper and the target. Note that this loss function can adjust all the items $(p^1, p^2, d, v)$ of the grasp pose.

Besides, we apply regularization loss to ensure that the grasp pose does not change too much, preventing excessive deviation from stable grasp poses:
\begin{equation}
\begin{aligned}
L_{reg} =\Vert \Delta p^{1} \Vert_2 + \Vert \Delta p^{2} \Vert_2 +  \Vert \Delta d \Vert_2 +  \Vert \Delta v \Vert_2,
\label{eq:loss_reg}
\end{aligned}
\end{equation}
where
$\Delta *$ denotes the grasp pose offset during the refinement process.

Among all the losses above, $L_{anti}$ and $L_{touch}$ fine-tune the grasp poses by directly changing the projected grasp points $(p^1$, $p^2)$ while the effect of $L_{collision}$ and $L_{reg}$ is performed by changing all the items of the grasp pose $(p^1, p^2, d, v)$. Our final refinement loss is defined as:
\begin{equation}
L_{r} = \lambda_{1} L_{anti}+\lambda_{2} L_{collision} + \lambda_{3} L_{touch} + \lambda_{4} L_{reg},
\label{eq:loss_refine}
\end{equation}
where $\lambda_1$, $\lambda_2$, $\lambda_3$ and $\lambda_4$ are the trade-off parameters to balance each loss. We set  $\lambda_1$, $\lambda_2$, $\lambda_3$ and $\lambda_4$ as 100, 10, 20 and 200 respectively in practice. During refinement, the weights of the DIF-Decoder and the latent shape code $z_t$ are fixed and $L_r$ is back-propagated through the DIF-Decoder to the grasp pose $g_t$, so that $g_t$ can be refined to a more stable grasp using Adam algorithm \cite{kingma2014adam}.

\section{Experiments}
\label{sec:experiments}

In this section, we first introduce the implementation details, along with the dataset and evaluation metrics we adopt.
Then we conduct experiments to verify the effectiveness of each part of TransGrasp in Sec.~\ref{sec:ablation} and make comparisons with other methods in Sec.~\ref{sec:comparative}.
Finally, real robot experiments are conducted to explore the practical application of our proposed method in Sec.~\ref{sec:robot_exp}.

\noindent
\textbf{Implementation Details:}
% \textcolor[rgb]{1,0,0}{
DIF-Decoder is trained following \cite{zheng2021deep}. For PointNet-like \cite{qi2017pointnet} shape encoder, we use the Adam optimizer \cite{kingma2014adam} with an initial learning rate of 0.0001 for training which takes 30 epochs for each category.
For the trade-off between the transfer success rate and optimization speed, we set the refine iteration number as 10 in practice.

\noindent
\textbf{Dataset:}
To evaluate our proposed TransGrasp, we select three household categories of objects including 100 mugs, 218 bottles and 107 bowls from ShapeNetCore \cite{shapenet2015}.
Meanwhile, considering the various shapes of objects within a class, we introduce data augmentation which enables the learned categorical shape latent space to cover possible object instances of a class.
For instance, mugs of different shapes and sizes are generated from the existing mug models by randomly enlarging or shrinking their top or bottom. Please refer to the supplementary for augmentation examples.
The training/test distribution of augmented data is 765/135 for mug, 940/150 for bottle and 801/162 for bowl, respectively.
We use only one object instance in each class of objects as the source object which are labeled hundreds of possible grasp poses (802 for mug, 1000 for both bottle and bowl, respectively).
From this labeled source object, its grasp poses will be transferred to other target objects of the same class.
To train Shape Encoder and 6D Object Pose Estimator, We render partial point clouds from 100 random views for each model in training set using Pytorch3D \cite{ravi2020pytorch3d}.

\noindent
\textbf{Evaluation Metrics:}
In this section, we evaluate the effectiveness of grasp pose estimation for the target objects of each category. As there are no ground truth grasp pose annotations on the target objects for comparison, we utilize IsaacGym \cite{makoviychuk2021isaac} to build a parallel-jaw simulation platform where the success rate of grasp transfer is replaced by the success rate of grasp.
Our simulation platform is shown in Fig.~\ref{fig:result_sim}(a).
All simulations are done using a Franka Panda manipulator.
Different from other grasp generation methods that are interested in the score of one best grasp among all predicted grasps, we focus on every transferred grasp pose and aim to evaluate the effect of our grasp pose transfer. Therefore, for each instance, we count the ratio of successful grasps among all transferred grasps as transfer success rate. And then the transfer success rate for one category is calculated as the average of all its instances' ones:
\begin{equation}
s_{trans} = \frac{1}{M}\sum_{i}^{M}  \frac{N^{suc}_{i}}{N^{all}_{i}},
\label{eq:successrate}
\end{equation}
where $M$ denotes the number of object instances while $N^{suc}$ and $N^{all}$ denote the number of successful grasps and total grasps respectively. Specifically, the platform consists of free-floating grippers and objects with gravity, and the grippers are placed according to their grasp poses. A grasp is considered as successful one if the object is held for more than 15s.

\subsection{Ablation Study}
\label{sec:ablation}
To verify the effectiveness of each part of TransGrasp and avoid other influencing factors, we first conduct the experiments using ground-truth 6D poses and scales of objects. Specifically, the following four experiments are conducted.
\textbf{\emph{Direct Mapping:}} The grasp points are directly obtained from the source labeled model and then scaled according to the size of target model while both depth $d$ and approaching vector $v$ remain unchanged.
\textbf{\emph{Direct Mapping w/ Refine:}} In addition to \emph{Direct Mapping}, it contains our grasp refinement module that fine-tunes grasp poses by changing grasp points ($p^1$, $p^2$), depth $d$ and approaching vector $v$.
\textbf{\emph{Grasp Transfer:}} The grasp points of target model are transferred from the source labeled model by our method and also scaled according to the size of target model while both depth $d$ and approaching vector $v$ remain unchanged.
\textbf{\emph{Grasp Transfer w/ Refine:}} With both grasp transfer and refinement module, it's the complete version of our proposed method.
The quantitative results are shown in Table~\ref{tab:com2self}. 

\begin{table}[t]
	\renewcommand\arraystretch{0.9}
	\begin{center}
	\setlength{\belowcaptionskip}{-0.0em}
	\caption{Performance of grasp transfer for each category}
	\label{tab:com2self}
		\setlength{\tabcolsep}{1em}
		{
			\begin{tabular}{c ccc c}
				\toprule
				\multirow{2}{*}{Method}  &  \multicolumn{4}{c}{Transfer Success Rate (\%)} \\
				\cline{2-5}
				& Mug  & Bottle  & Bowl & Average\\
				\midrule
				Direct Mapping          & 60.68 & 79.81 & 33.27 & 57.92\\
				Direct Mapping w/ Refine  & 59.26 & 79.86 & 34.83 & 57.98\\
				Grasp Transfer            & 82.67 & 86.93 & 46.24 & 71.95\\
				Grasp Transfer w/ Refine        & \textbf{87.77} & \textbf{87.24} & \textbf{61.52} & \textbf{78.84}\\
				\bottomrule
			\end{tabular}
		}
	\end{center}
\end{table}

Since grasp points after direct mapping may not be on the surface of target object, which provides a poor initialization for refinement, refinement after direct mapping barely has improvement (57.92\% to 57.98\% in average). Compared to \emph{Direct Mapping}, \emph{Grasp Transfer} greatly increases grasp success rate with 14.03\% for the average success rate. This is because through deformation, the transferred grasp points can adapt to object instances with different shapes, which largely avoids the collision of the gripper with the object. It benefits from the categorical shape space learned by TransGrasp which establishes dense correspondences. Besides, \emph{Grasp Transfer w/ Refine} with the proposed grasp pose refinement module further increases the grasp transfer success rates with 6.89\% for the average success rate, indicating that our grasp pose refinement module effectively optimizes the transferred grasp poses and improves the stability of grasping.
It is worth noting that each method has a relatively low grasp success rate for the bowl. The first reason is that about 28\% of the bowl models are single-layer meshes which are not suitable for grasping.
The second reason is that the grasp poses of the bowl are always surrounding the rim of bowl, far away from the center of mass of the bowl, which makes the external force insufficient to produce friction that can resist the gravity of the bowl.

\begin{figure}[t]
    \centering
	\includegraphics[width=0.84\linewidth]{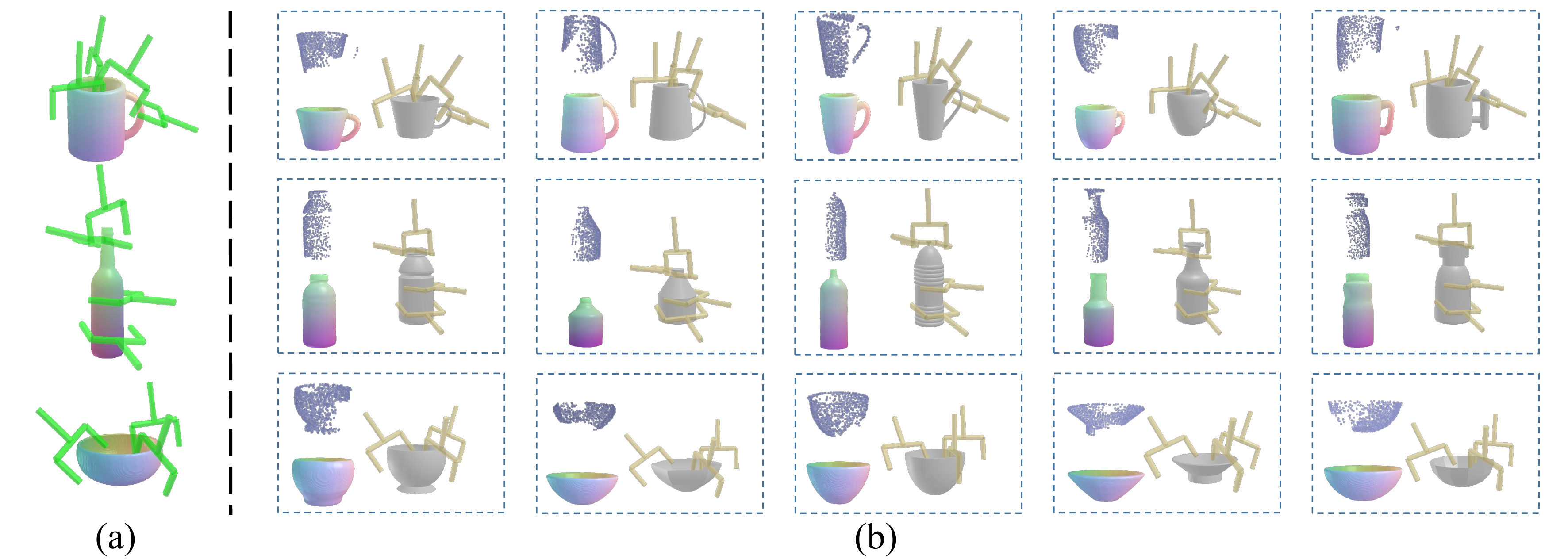}
	\setlength{\abovecaptionskip}{-0.4em}
	\caption{
Qualitative results of three categories. (a) shows source labeled models and some grasp pose annotations. (b) illustrates the transferred grasp poses. For every instance represented inside a dashed box, we show its input partial point cloud (upper left), reconstructed model (bottom left) and transferred grasp poses (right)
	}
	\label{fig:result}
\end{figure}

\begin{figure}[t]
    \centering
% 	\begin{center}
		\includegraphics[width=0.9\linewidth]{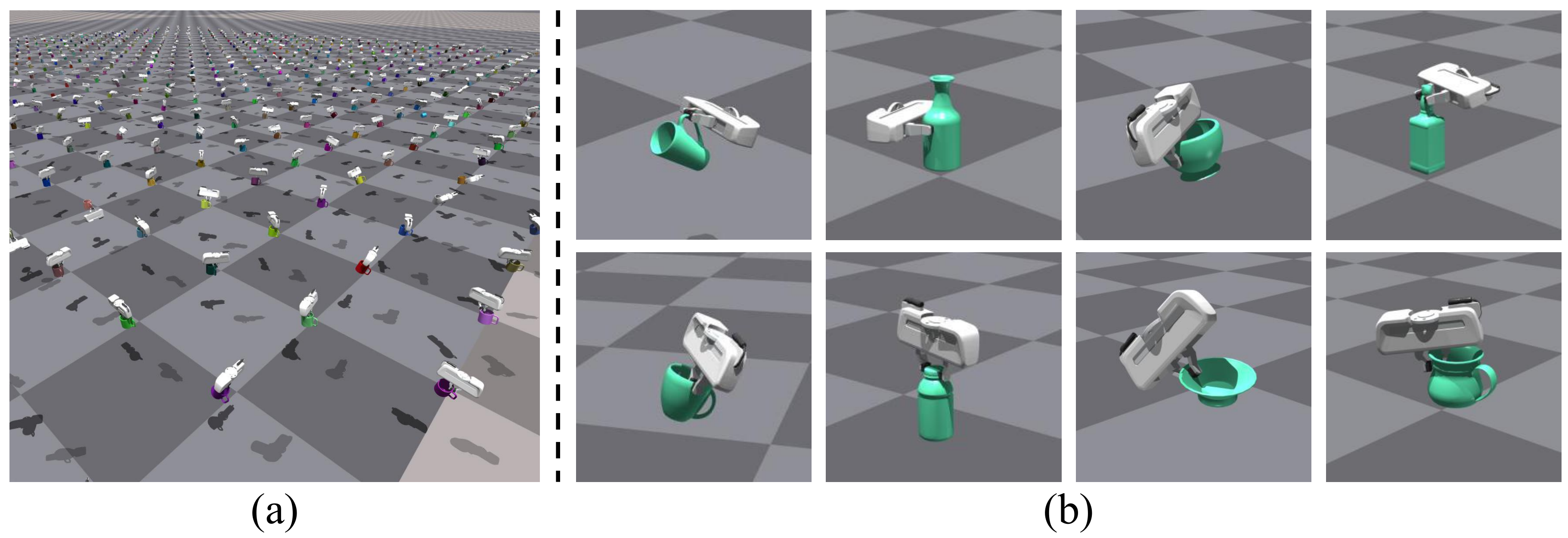}
% 	\end{center}
% 	\setlength{\abovecaptionskip}{-0.7em}
	\setlength{\abovecaptionskip}{-0.1em}
	\caption{
		Simulation platform and some grasp results
	}
	\label{fig:result_sim}
\end{figure}

Figure~\ref{fig:result} shows the qualitative results of the transferred grasps after refinement on the test set of each category. We only show a few grasps for clarity. It can be observed that our method can efficiently estimate high-quality grasps even if the input data is incomplete observed point cloud of unseen instance. Figure~\ref{fig:result_sim}(b) shows the grasp results on the simulation platform \cite{makoviychuk2021isaac}. It can be seen that the grippers can grasp objects stably using our transferred grasp poses, even for those objects with slender handle of the mug and thin edge of the bowl.

\subsection{Comparisons with Other Methods in Simulation}
\label{sec:comparative}
The above ablation experiments verify the effectiveness of the proposed TransGrasp, we continue to compare our method with others in this section.
Among the grasping methods, GPD \cite{ten2017grasp} and 6-DOF GraspNet \cite{mousavian20196} have similar grippers and object samples as ours, and the two methods are also representative grasping methods at present. Therefore, we make comparison with them. The major difference from the above two methods is that TransGrasp only labels one instance in grasp pose estimation of an entire category of objects, while GPD and 6-DOF GraspNet require grasp pose annotations for every object.

\begin{table}[t]
	\renewcommand\arraystretch{0.9}
	\begin{center}
	\setlength{\belowcaptionskip}{-0.0em}
	\caption{Quantitative comparison of grasp success rate and inference time}
	\label{tab:com2other}
		\setlength{\tabcolsep}{0.4em}
		{
			\begin{tabular}{c ccc c c}
				
				\toprule
				\multirow{2}{*}{Method}  &  \multicolumn{4}{c}{Grasp Success Rate (\%)} & Inference Time (s)\\
				\cline{2-5}
				& Mug  & Bottle  & Bowl & Average & (NVIDIA 1080 GPU)\\
				\midrule
				GPD \cite{ten2017grasp}                & 34.81 & 74.00 & 44.44 & 51.08 & 1.14\\
				6-DOF GraspNet  \cite{mousavian20196}    & 76.30 & 78.00 & \textbf{75.93} & 76.74 & 1.82 \\
				TransGrasp (Ours)      & \textbf{86.67}  & \textbf{88.00} & 72.84  & \textbf{82.50} & \textbf{0.53} \\
				\bottomrule
			\end{tabular}
		}
	\end{center}
\end{table}

In this experiment, we evaluate grasp success rate on test set of three categories: \emph{mug}, \emph{bottle} and \emph{bowl}. We employ \cite{tian2020shape} as the 6D Object Pose Estimator to estimate 6D pose of the input point cloud of a target object, as shown in the left of Fig.~\ref{fig:overview},
by which the grasp pose is transformed from Object Coordinates to Camera Coordinates for the next step.
We choose the grasp pose with minimal refinement loss as the final one. For GPD and 6-DOF GraspNet, their final grasp poses are chosen according to the grasp scores computed by their own evaluation networks.
To be fair, all the methods generate 100 grasp samples for selection.

Table~\ref{tab:com2other} shows the quantitative results of the three methods. Compared with GPD and 6-DOF GraspNet, TransGrasp obtains the highest success rate for the mug and bottle category (88.67\% and 88.00\%), and achieves comparable performance with 6-DOF GraspNet for the bowl category although our method only labels one instance of an entire category of objects. 
One dominant factor is that our method reconstructs the complete shapes of objects,
rather than directly estimate grasp poses from partial input point clouds like GPD and 6-DOF GraspNet, which introduces more feasible grasps. And benefiting from the reconstructed complete shape, our refinement module can avoid the interpenetration between gripper and unseen parts of object.
Furthermore, even our method contains the refinement module, it takes the least inference time with 0.53s among these methods, which satisfies the requirement of grasping household objects.

\subsection{
% \textcolor[rgb]{1,0,0}{
Real Robot Experiments
% }
}
\label{sec:robot_exp}

\begin{table}[t]
	\renewcommand\arraystretch{0.9}
	\begin{center}
	\setlength{\belowcaptionskip}{-0.0em}
	\caption{Quantitative comparison of grasp success rate in real robot experiments}
	\label{tab:com2other_robot}
		{
			\begin{tabular}{c ccc c}
				
				\toprule
				Method & Mug  & Bottle  & Bowl & Average\\
				% \hline
				\midrule
				
				GPD \cite{ten2017grasp}  & 7/25(28\%) & 19/25(76\%) & 11/25(44\%) & 37/75(49.3\%) \\
				6-DOF GraspNet\cite{mousavian20196} & 16/25(64\%) & 17/25(68\%) & \textbf{19/25(76\%)} & 52/75(69.3\%) \\
				TransGrasp (Ours) & \textbf{19/25(76\%)}  & \textbf{21/25(84\%)} & \textbf{19/25(76\%)}  & \textbf{59/75(78.7\%)} \\
				\bottomrule
			\end{tabular}
		}
	\end{center}
\end{table}

\begin{figure}[t]
    \centering
	\includegraphics[width=0.9\linewidth]{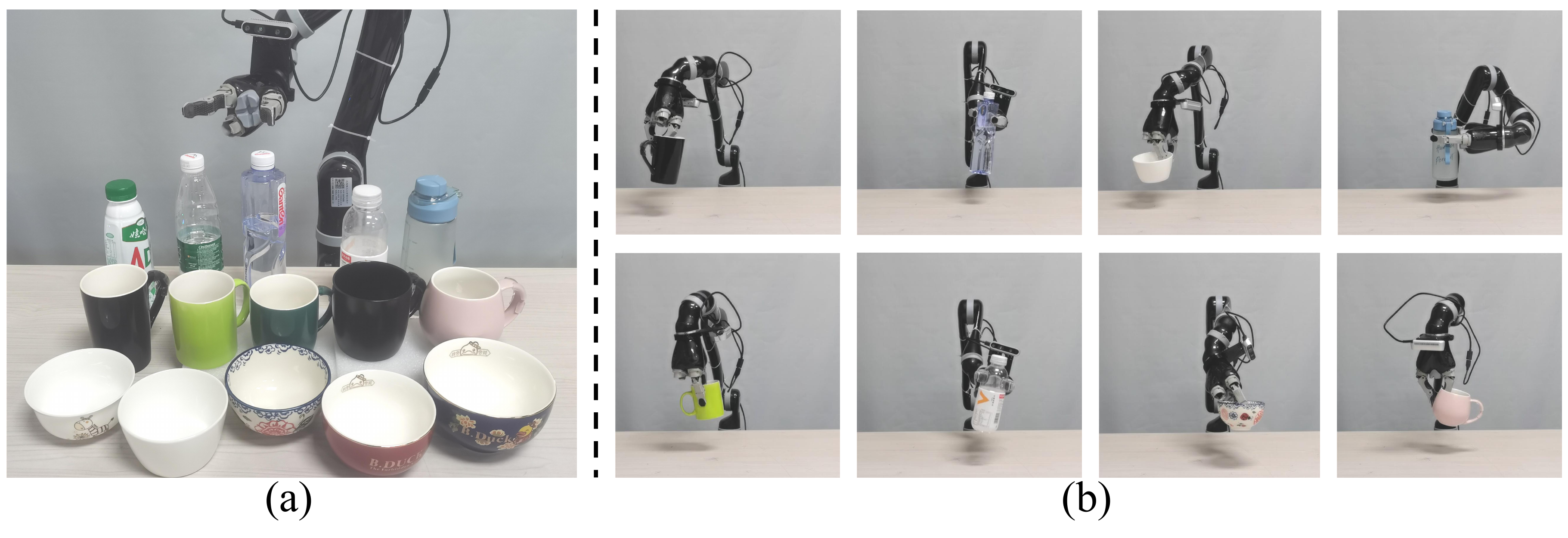}
	\setlength{\abovecaptionskip}{-0.6em}
	\caption{
		Real-world experiments setting and some practical results
	}
	\label{fig:robotEXP}
\end{figure}
To verify the practical application of our method, we further conduct grasping experiments on a real robot. All experiments are done using a JACO manipulator with an Intel RealSense D435 mounted on its gripper. We choose 5 objects per category with weights from 135.6g to 436.2g for grasping. Figure~\ref{fig:robotEXP}(a) shows the hardware setup and object test set. We make comparison with GPD and 6-DOF GraspNet in our experiments. During the grasping process, we first classify and segment each instance in the scene using off-the-shelf instance segmentation method \cite{wang2020solo} and then estimate 6D object pose of each instance using \cite{tian2020shape}.
We transfer the grasps with the models trained on our synthetic dataset and test in real scenario directly for our method. For GPD and 6-DOF GraspNet, we use the trained models given by authors to generate grasp poses.
After grasps are generated or transferred, we choose the grasp using the grasp selection strategy consistent with the simulation experiments in Sec.~\ref{sec:comparative}. If the gripper can lift the object, the grasp is marked as success.
We run 5 trials per object and totally 25 trials for each category. As shown in Table~\ref{tab:com2other_robot}, our method outperforms GPD and 6-DOF GraspNet on success rate for mug and bottle, and  achieves the same performance with 6-DOF GraspNet for bowl. Figure~\ref{fig:robotEXP}(b) shows some of our grasping results in the real-world setting. It can be observed that the grippers can successfully grasp objects with different shapes in real scenes, which indicates that our method can be applied to practical applications. More experiment results can be seen in the supplementary material.

\section{Limitations and Future Work}
\begin{figure}[h]
    \centering
	\includegraphics[width=0.85\linewidth]{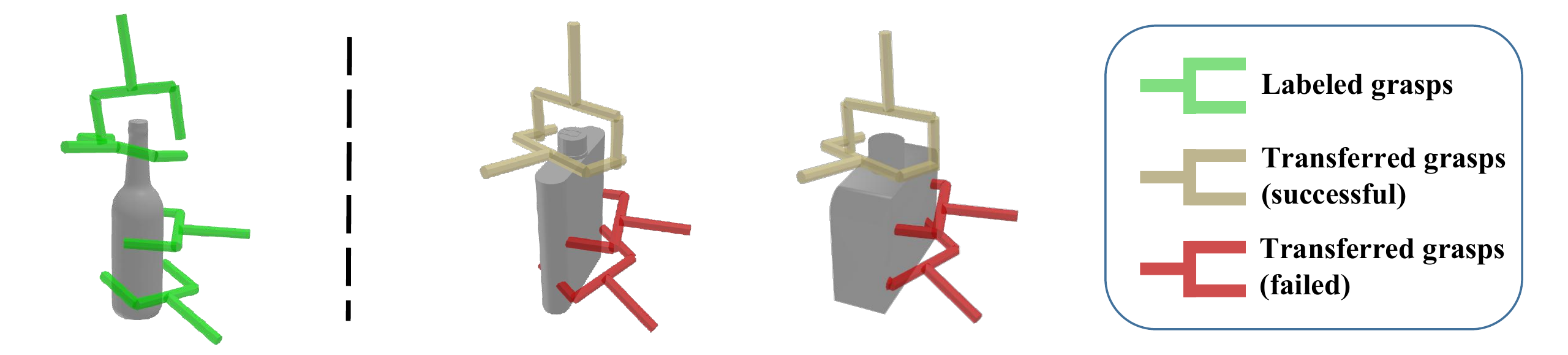}
	\setlength{\abovecaptionskip}{-0.3em}
	\caption{
Grasp Transfer with large shape variances for the bottle category
	}
	\label{fig:fail_case}
\end{figure}

Although the experiments have demonstrated the effectiveness of our method, there is still a limitation to be addressed. As shown in Fig.~\ref{fig:fail_case}, the grasps are difficult to be transferred across objects with large shape variances. When the grasps on a cylindrical bottle are transferred to a flat bottle or a square bottle, some of the transferred grasps (in red) penetrate the object. We can further improve the transfer success rate by classifying these objects into fine-grained subclass in the future.
\section{Conclusion}
In this work, we present a category-level grasp pose estimation method to predict grasp poses of a category of objects by labeling only one object instance, which does not rely on a large amount of grasp annotations or exact 3D object models available beforehand. We firstly establish dense shape correspondence among an entire category of objects to transfer the grasp poses from the one source object to other instances, and then build a grasp pose refinement module to improve performance after grasp transfer.
Sufficient experiments show that our estimated grasp poses can align well across the instances of one category and achieve high-quality grasp performance compared with existing methods.
\\

\noindent \textbf{Acknowledgments.}
This research was supported by the National Natural Science Foundation of China (No.U1708263 and No.61873046).

\bibliographystyle{splncs04}
\bibliography{egbib}

\clearpage

\appendix
\begin{center}
\Large{\bfseries{\boldmath{
TransGrasp: Grasp Pose Estimation of a\\
Category of Objects by Transferring Grasps\\
from Only One Labeled Instance\\
(Supplementary Material)}}}
\end{center}
\begin{table}[!h]
	\begin{center}
	\caption{Detailed results of real robot experiments in the Table~\ref{tab:com2other_robot} of our main paper
	% \\\dui : successful trial, \cuo : failed trial.
	}
	\label{tab:detailed_results}
	\begin{tabular}{c|c|ccccc|ccccc|ccccc}
	\toprule[1.3pt]
	\multirow{2}{*}{Category} & \multirow{2}{*}{Instances} & \multicolumn{5}{c|}{GPD\cite{ten2017grasp}}     & \multicolumn{5}{c|}{6-DOF GraspNet\cite{mousavian20196}}     & \multicolumn{5}{c}{TransGrasp}     \\
							\cline{3-17}
							%   &                            & 
							%   \multicolumn{15}{c}{trials Number} \\
							%   \multicolumn{5}{c|}{trials}  & \multicolumn{5}{c|}{trials}  & \multicolumn{5}{c}{trials}  \\ 
							%   \cline{3-17} 
	 &   & \#1 & \#2 & \#3 & \#4 & \#5 & \#1 & \#2 & \#3 & \#4 & \#5 & \#1 & \#2 & \#3 & \#4 & \#5 \\ 
	\midrule[1.1pt]
	\multirow{5}{*}[-22pt]{Mug}     
	& \raisebox{-.5\height}{\includegraphics{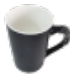}}&\dui & \cuo & \dui & \cuo & \cuo & \dui & \dui & \cuo & \dui & \cuo & \cuo & \dui & \dui & \dui & \dui \\
	& \raisebox{-.5\height}{\includegraphics{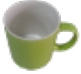}}   &  \cuo   &  \cuo   &  \cuo   &  \cuo   &  \cuo    &  \cuo   &  \dui   & \cuo    &  \cuo   &   \dui   &  \cuo   &   \cuo  &  \dui   &  \cuo   &  \cuo     \\
	& \raisebox{-.5\height}{\includegraphics{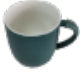}}   &  \cuo   &   \dui  &  \cuo  &  \cuo   &  \cuo   &   \dui   &  \dui   &  \dui   &  \dui   &  \cuo    &  \cuo   &   \dui  &  \dui   & \dui    &  \dui \\
	& \raisebox{-.5\height}{\includegraphics{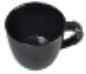}}   &  \cuo    &   \dui  &  \dui   &  \dui   &  \dui    &  \dui   &  \cuo    &   \cuo   &  \dui   &  \dui    &  \dui   &  \dui   &  \dui   &  \dui   &  \dui   \\
	& \raisebox{-.5\height}{\includegraphics{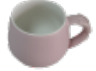}}   &  \cuo    &  \cuo    &  \cuo    &  \cuo    &  \cuo   &  \cuo    &   \dui  &  \dui   &  \dui   &  \dui    &  \dui   &  \dui   &   \dui  &  \dui   &  \dui   \\
	\cline{2-17}
	& \multirow{1}{*}[-1pt]{ Success Rate }  &  \multicolumn{5}{c|}{7 / 25}     &  \multicolumn{5}{c|}{16 / 25}    &  \multicolumn{5}{c}{\textbf{19 / 25}}   \\
	\midrule
	\multirow{5}{*}[-30pt]{Bottle}   
	& \raisebox{-.5\height}{\includegraphics{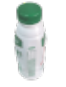}}& \dui & \dui & \dui & \dui & \dui & \cuo & \dui & \cuo & \dui & \dui & \dui & \dui & \dui & \cuo & \dui \\
	& \raisebox{-.5\height}{\includegraphics{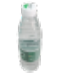}}& \cuo & \cuo & \cuo & \dui & \cuo & \cuo & \cuo & \dui & \cuo & \cuo & \cuo & \dui & \cuo & \dui & \dui \\
	& \raisebox{-.5\height}{\includegraphics{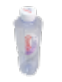}}& \dui & \dui & \dui & \dui & \dui & \dui & \cuo & \dui & \dui & \cuo & \dui & \dui & \dui & \dui & \dui \\
	& \raisebox{-.5\height}{\includegraphics{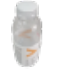}}& \cuo & \dui & \dui & \dui & \dui & \dui & \dui & \dui & \dui & \dui & \dui & \dui & \dui & \dui & \dui \\ 
	& \raisebox{-.5\height}{\includegraphics{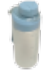}}& \cuo & \dui & \dui & \dui & \dui & \dui & \dui & \dui & \dui & \dui & \dui & \dui & \dui & \dui & \cuo \\ 
	\cline{2-17}
	& \multirow{1}{*}[-1pt]{ Success Rate }  &  \multicolumn{5}{c|}{19 / 25}     &  \multicolumn{5}{c|}{17 / 25}    &  \multicolumn{5}{c}{\textbf{21 / 25}}   \\
	\midrule
	\multirow{5}{*}[-15pt]{Bowl}
	& \raisebox{-.5\height}{\includegraphics{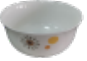}}  &  \cuo   &  \cuo   &  \cuo   &  \dui   &   \cuo    &  \cuo   &  \cuo   & \dui    & \dui    &   \dui   &   \dui  &  \dui   &  \dui   &  \dui   &  \dui   \\
	& \raisebox{-.5\height}{\includegraphics{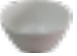}}&   \cuo    &  \dui   &  \cuo   & \dui    &    \cuo   &   \dui  &  \dui   &  \dui   &   \dui  &   \dui    &  \dui   &   \dui  &   \cuo  &  \dui   &  \dui   \\
	& \raisebox{-.5\height}{\includegraphics{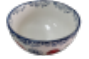}}  &  \dui     &  \dui   &  \cuo   &  \cuo   &   \cuo    &  \dui   &   \dui  &  \dui   & \dui    & \cuo    &   \cuo   &  \dui   &  \dui   &   \cuo  &   \dui  \\
	& \raisebox{-.5\height}{\includegraphics{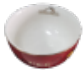}}  &    \cuo  &  \dui   &   \dui  &  \dui   &   \dui   &   \dui  &  \dui   &  \cuo   &  \dui   &  \dui   &  \cuo   &  \dui   &  \dui   &   \cuo  &   \dui  \\
	& \raisebox{-.5\height}{\includegraphics{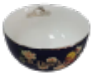}}  &   \dui  &  \dui   &  \cuo   &  \cuo   &  \cuo   &   \cuo  &   \dui   &  \dui   &  \cuo   &  \dui   &  \dui   &  \dui   & \cuo    &   \dui  &   \dui  \\
	\cline{2-17}
	& \multirow{1}{*}[-1pt]{ Success Rate }  &  \multicolumn{5}{c|}{11 / 25}     &  \multicolumn{5}{c|}{\textbf{19 / 25}}    &  \multicolumn{5}{c}{\textbf{19 / 25}}   \\
	\midrule
	\multicolumn{2}{c|}{ Average Success Rate }  &  \multicolumn{5}{c|}{37 / 75}     &  \multicolumn{5}{c|}{52 / 75}    & 
	\multicolumn{5}{c}{\textbf{59 / 75}} \\
	\bottomrule[1.3pt]
	\end{tabular}
	\end{center}
	\vspace{-1em}
	\end{table}
\clearpage
\section{Detailed Results of Real Robot Experiments}
Table~\ref{tab:detailed_results}, where \dui and \cuo denote successful and failed trials, respectively, shows the detailed results of the real robot experiments in Table~\ref{tab:com2other_robot} of our main paper. For each method, we performed 5 trials per object and totally 25 trials per category. 
	Our method outperforms GPD\cite{ten2017grasp} and 6-DOF GraspNet\cite{mousavian20196} for the mug and bottle category, and achieves the same performance as 6-DOF GraspNet for the bowl category. 
	% Note that for the transparent bottle in the seventh row whose point cloud contains background points, the success rate of our method is significantly higher than that of other methods (3/5 for ours \vs 1/5 for both GPD and 6 DOF-GraspNet). This is because our method can reduce the influence of background noise from point clouds on grasp generation by reconstruction complete shape of the target, while GPD and 6-DOF GraspNet that generate grasps directly from point clouds cannot.
	It's noteworthy that for the transparent bottle, the success rate of our method is significantly higher than that of GPD and 6-DOF GraspNet. Owing to the shape reconstruction for the target object, our method can easily reduce the influence of background points on grasp estimation, unlike GPD and 6-DOF GraspNet that generate grasps directly from point clouds. 
	
	% We show the process of grasping for our method in the supplementary video. It can be observed that the grippers can grasp objects stably for different instances of the three categories, which verifies the effectiveness of our method in practical applications.
	
	\section{Objects of Various Shapes in Each Category}
	\begin{figure}[!h]
	\vspace{-2em}
		\centering
			\includegraphics[width=\linewidth]{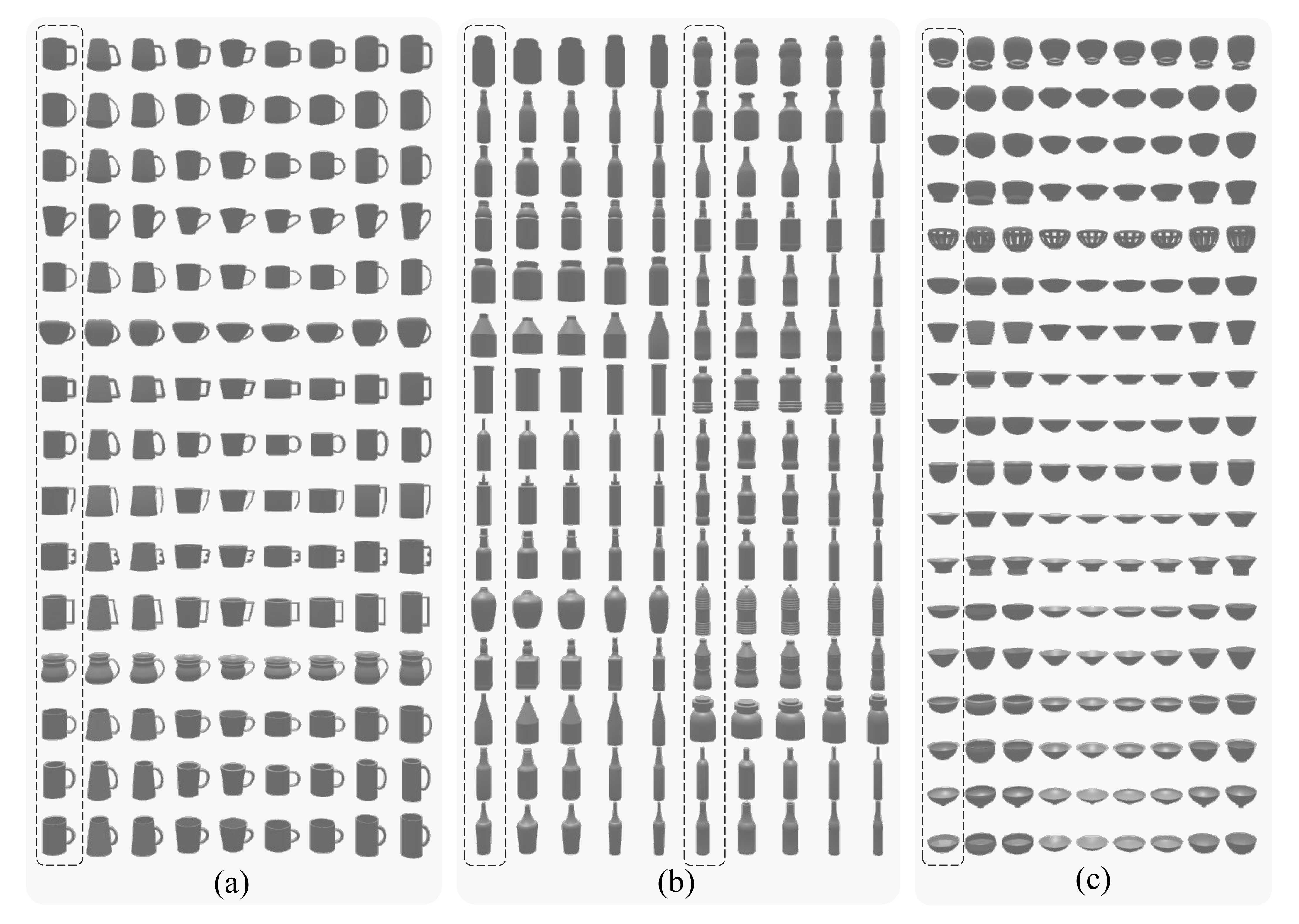}
		\caption{
			Objects of various shapes in the test sets of each category, where the models in the dashed box are the original models selected from ShapeNetCore \cite{shapenet2015} and the others are the models after augmentation
		}
		\label{fig:aug_model_eval}
	\end{figure}
	To increase the diversity of shapes in the dataset, we introduce a data augmentation strategy by deforming the models selected from ShapNetCore \cite{shapenet2015} as shown in Fig.~\ref{fig:aug_model_eval}. Specifically, for each model of these three categories, its height is first increased or decreased and then it is normalized to unit space to produce new objects with different ratios of length, width and height. To further increase the diversity of the shapes for the mug and bowl category, we additionally perform deformations by enlarging and shrinking the mouths of them. As shown in Fig.~\ref{fig:aug_model_eval}, There are various shapes of objects in our test set, which are enough to evaluate the effectiveness of our proposed grasp pose estimation method.
	
	\section{Effect of Grasp Pose Refinement}
	\begin{figure}[!h]
	\vspace{-2em}
		\begin{center}
			\includegraphics[width=0.95\linewidth]{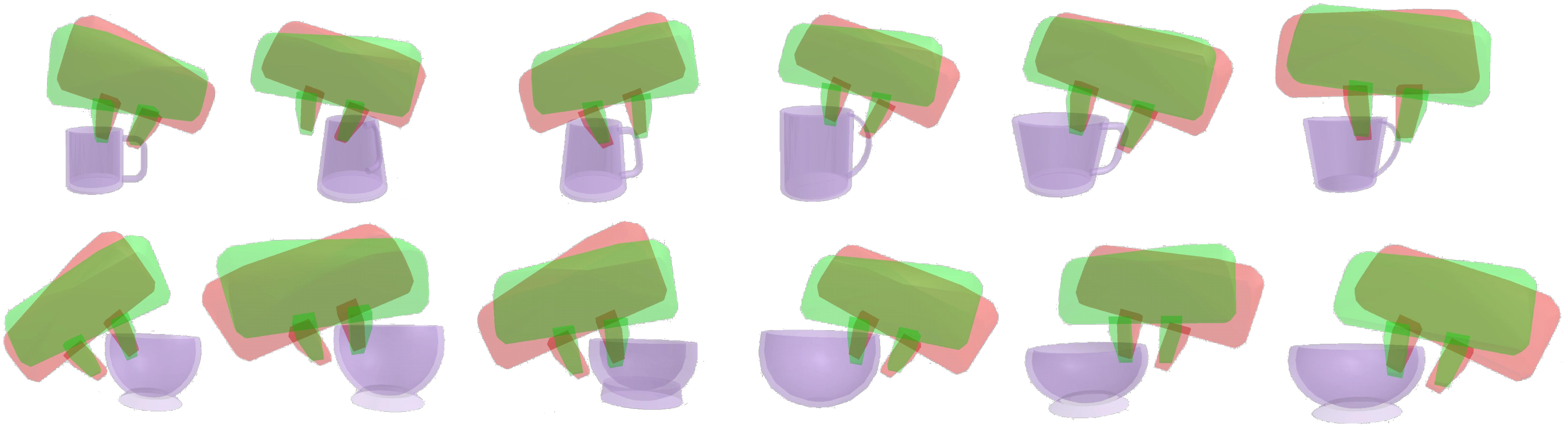}
		\end{center}
		\setlength{\abovecaptionskip}{-1.4ex}
		\caption{
			% 
	% 		Effect of $L_{anti}$.
	\textcolor[rgb]{0.871,0.357,0.376}{Red} grippers are % represent the grasp poses 
	not refined with the antipodal refinement loss $L_{anti}$ and \textcolor[rgb]{0.604,0.941,0.604}{green} grippers are refined with $L_{anti}$
		}
		\label{fig:refine_anti}
		\vspace{-1ex}
	\end{figure}
	
	\begin{figure}[!h]
	\vspace{-2em}
		\begin{center}
			\includegraphics[width=0.95\linewidth]{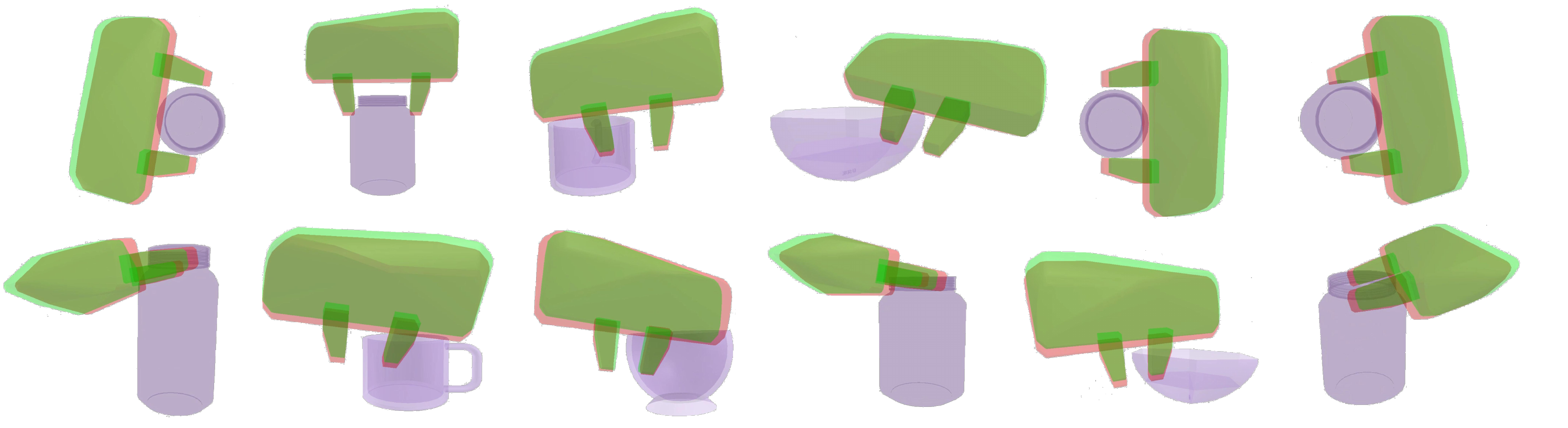}
		\end{center}
		\setlength{\abovecaptionskip}{-1.4ex}
		\caption{
			% 
	% 		Effect of $L_{anti}$.
	% 		\textcolor[rgb]{0.604,0.941,0.604}{Green} and \textcolor[rgb]{0.871,0.357,0.376}{red} grippers represent the grasp poses refined with and without the collision avoidance refinement loss $L_{collision}$.
	\textcolor[rgb]{0.871,0.357,0.376}{Red} grippers are not refined with the collision avoidance refinement loss $L_{collision}$ and \textcolor[rgb]{0.604,0.941,0.604}{green} grippers are refined with $L_{collision}$
		}
		\label{fig:refine_collision}
		\vspace{-1ex}
	\end{figure}
\begin{figure}[h]
		\begin{center}
			\includegraphics[width=0.9\linewidth]{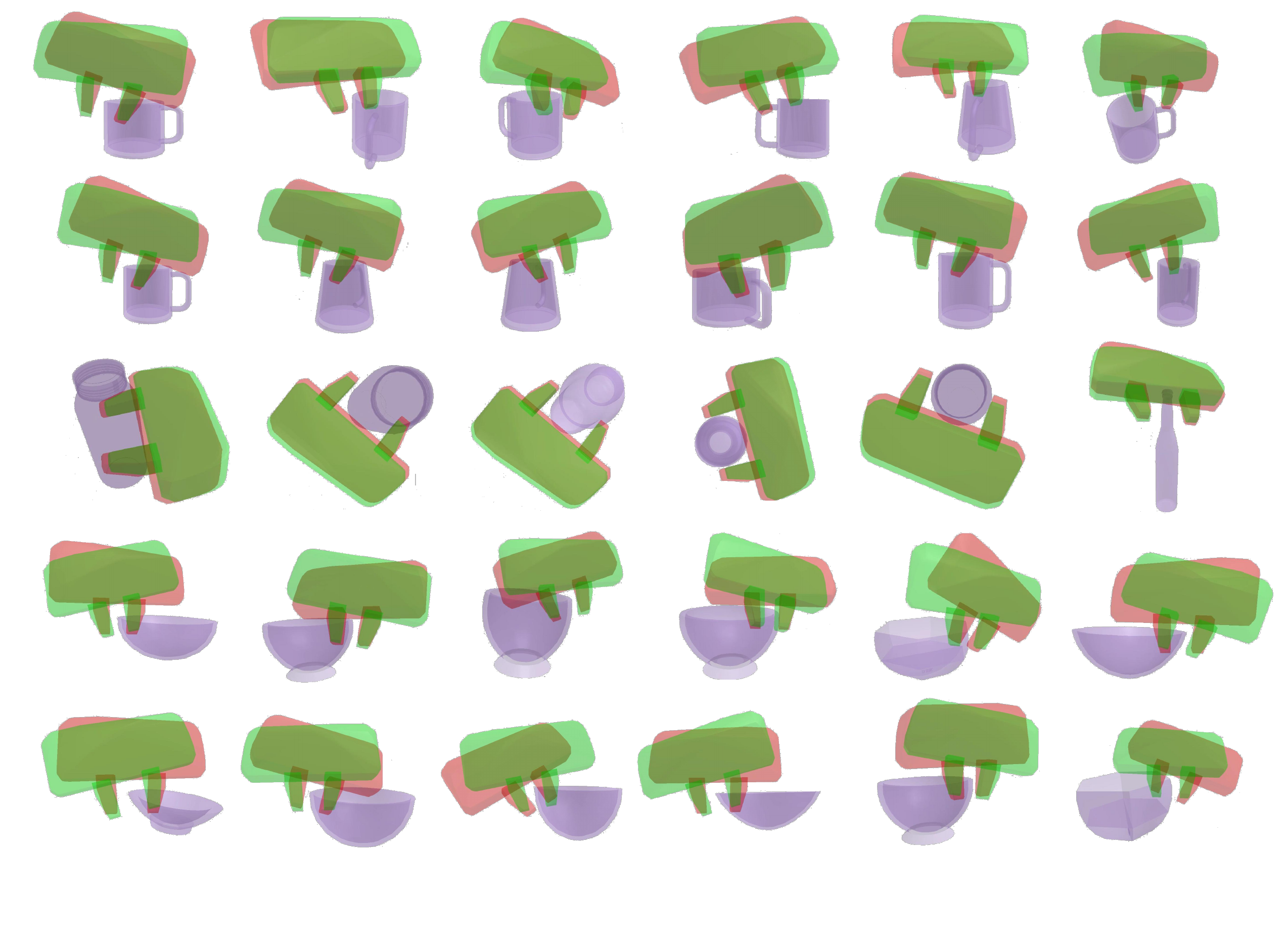}
		\end{center}
		\setlength{\abovecaptionskip}{-3em}
		\caption{
			Qualitative results of the proposed grasp pose refinement module. \textcolor[rgb]{0.871,0.357,0.376}{Red}  grippers represent the grasp poses before refinement and  \textcolor[rgb]{0.604,0.941,0.604}{green} grippers represent the grasp poses after refinement
	% 		\textcolor[rgb]{0.871,0.357,0.376}{Red} and \textcolor[rgb]{0.604,0.941,0.604}{green} grippers represent the grasp poses before and after refinement respectively.
		}
		\label{fig:refine_all}
	% 	\vspace{-2em}
	\end{figure}

	We illustrate the contribution of two main loss functions defined in our main paper in Fig.~\ref{fig:refine_anti} and Fig.~\ref{fig:refine_collision}. Clearly, the antipodal loss $L_{anti}$ that encourages the grasping points to satisfy the antipodal principle guides the gripper to achieve a more stable grasp shown in Fig.~\ref{fig:refine_anti}. Moreover, the green gripper refined with $L_{collision}$ can keep a safer distance from the object than the red one not refined with $L_{collision}$ in Fig.~\ref{fig:refine_collision}, avoiding interpenetration between gripper and object. The two refinement loss functions, together with $L_{touch}$ forcing grasping points on the object surface and $L_{reg}$ avoiding grasp pose beyond the limit, jointly ensure successful grasp. More examples about the grasp poses before and after refinement are shown in Fig.~\ref{fig:refine_all}. 
	
	\section{Grasp Pose Representation}
	To connect grasp pose $[R, t]$ with grasping points on the object surface and facilitate the optimization of pose parameters, we also design a new grasp pose representation as $g=(p^{1}, p^{2}, d, v)$ where $(p^{1}, p^{2})$ denotes grasp points on the object surface, $d$ the approaching depth and $v$ the approaching vector. In this section, we describe in detail how these two representations are converted to each other. 
	
	\begin{figure}[!h]
		\vspace{-1em}
		\begin{center}
			\includegraphics[width=0.4\linewidth]{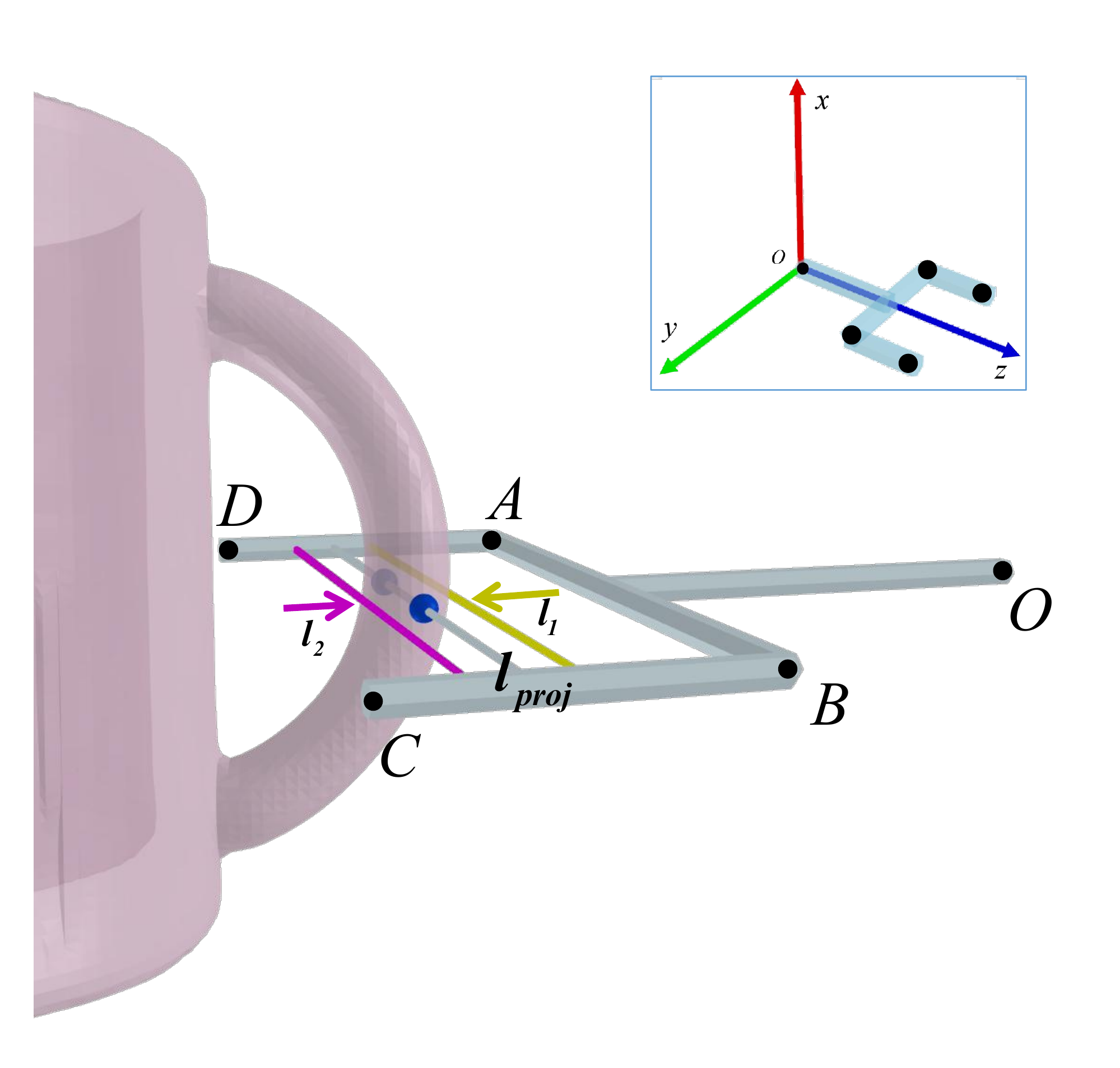}
		\end{center}
		\setlength{\abovecaptionskip}{-1.4ex}
		\caption{
			Process of obtaining grasp points on mug handle surface
		}
		\label{fig:Rt_2_g}
		\vspace{-1.5em}
	\end{figure}
	The conversion process from $[R, t]$ to $g$ is illustrated in Algorithm~\ref{algo_Rt_2_g}. We define 4 corner points on the gripper and transform these points to the Object Coordinates (denoted as $[A, B, C, D]$) using the gripper pose $[R, t]$ as shown in Fig.~\ref{fig:Rt_2_g}. To obtain the grasp points $(p^1, p^2)$ on the object surface, we firstly define two searching lines $l_1$ in \textcolor[rgb]{0.784,0.784,0.157}{yellow} and $l_2$ in \textcolor[rgb]{0.745,0.,0.745}{purple} which are parallel to each other with the starting positions coincided with $AB$ and $DC$ respectively. The two lines move towards each other and any line stops when contacting the object surface. After they both stop, their mid-line is then determined as the projection line $l_{proj}$.Finally, where the projection line $l_{proj}$ intersects the object surface is the grasp points $(p^{1}, p^{2})$. In addition, the approaching vector $v$ is easily derived from $[R, t]$ and the approaching depth $d$ is the distance from gripper origin point $O$ to $l_{proj}$. 
	
	In grasping, grasp pose representation $[R, t]$ is usually used to adjust the gripper, thus we need to convert the estimated grasp pose $g$ back to $[R, t]$ representation. The detailed conversion process is explained in Algorithm~\ref{algo_g_2_Rt}.
	\vspace{-1em}
	\begin{algorithm}[!h]
	\caption{Conversion from $[R, t]$ to $g$}
	\begin{algorithmic}[1]
	\State \textbf{Input:} gripper model $G$, grasp pose matrix $[R,t]$ and object model $M$
	\State $[O,A,B,C,D] \gets GetCornerPoints(G,R,t)$
	\State $i \gets 0$
	\Repeat
		\State $E \gets (A*(100-i) + D*i)/100$
		\State $F \gets (B*(100-i) + C*i)/100$
		\State $i \gets i+1$
	\Until{$IsContact(EF, M)$}
	\State $l_1 \gets EF$
	\State $i \gets 0$
	\Repeat
		\State $E \gets (D*(100-i) + A*i)/100$
		\State $F \gets (C*(100-i) + B*i)/100$
		\State $i \gets i+1$
	\Until{$IsContact(EF, M)$}
	\State $l_2 \gets EF$
	\State $l_{proj} \gets (l_1 + l_2) / 2$
	\State $p^1,p^2 \gets GetGraspPoints(l_{proj}, M)$
	\State $v \gets GetApproachVector(R)$
	\State $d \gets GetDistance(O, l_{proj})$
	% \EndFor
	\State \textbf{Output:} grasp pose $g=(p^1,p^2,d,v)$
	\end{algorithmic}
	\label{algo_Rt_2_g}
	\end{algorithm}
	\vspace{-3em}
	\begin{algorithm}[!h]
	\caption{Conversion from $g$ to $[R, t]$}
	\begin{algorithmic}[1]
	\State \textbf{Input:} grasp pose $g=(p^1,p^2,d,v)$
	% \State $y_{rot} \gets p^2 - p^1$
	\State $y_{rot} \gets (p^2 - p^1) / |p^2 - p^1|$
	\State $x_{rot} \gets y_{rot} \times v$
	\State $x_{rot} \gets x_{rot} / |x_{rot}|$
	\State $z_{rot} \gets x_{rot} \times y_{rot}$
	\State $R \gets Concatenate \left (x_{rot},y_{rot},z_{rot} \right )$
	\State $t \gets (p^1 + p^2) / 2 - z_{rot} * d$
	\State \textbf{Output:} grasp pose matrix $[R, t]$
	\end{algorithmic}
	\label{algo_g_2_Rt}
	\end{algorithm}	

\end{document}